\documentclass[runningheads]{llncs}

\usepackage{graphicx}
\usepackage{amsmath,amssymb,amsfonts}
\usepackage{cleveref}
\crefname{section}{Sect.}{Sect.}
\Crefname{section}{Section}{Sections}
\crefname{figure}{Fig.}{Fig.} 

\usepackage[colorinlistoftodos]{todonotes}

\usepackage{amsmath}
\usepackage{algpseudocode}
\usepackage{algorithm}
\algnewcommand\algorithmicforeach{\textbf{for each}}
\algdef{S}[FOR]{ForEach}[1]{\algorithmicforeach\ #1\ \algorithmicdo}

\usepackage{tikz}
\tikzset{>=latex}

\usepackage[acronym]{glossaries}
\usepackage{glossaries-extra}
\usepackage[super]{nth}
\setabbreviationstyle[acronym]{long-short}
\newacronym{ai}{AI}{Artificial Intelligence}
\newacronym{ml}{ML}{Machine Learning}
\newacronym{rl}{RL}{Reinforcement Learning}
\newacronym{dl}{DL}{Deep Learning}
\newacronym{ot}{OT}{Operational Technology}
\newacronym{cps}{CPS}{Cyber-Physical System}
\newacronym{plc}{PLC}{Programmable Logic Controller}
\newacronym{hil}{HiL}{Hardware-in-the-Loop}
\newacronym[plural=ics,firstplural=Industrial Control Systems (ICS)]{ics}{ICS}{Industrial Control System}

\begin{document}
\title{
  An Architecture for Deploying Reinforcement Learning in Industrial Environments%
  \thanks{Georg Schäfer and Stefan Huber are supported by the European Interreg Österreich-Bayern project AB292 KI-Net and the Christian Doppler Research Association. Reuf Kozlica is supported by the Lab for Intelligent Data Analytics Salzburg (IDA Lab) funded by Land Salzburg (WISS 2025) under project number 20102-F1901166-KZP.}
}
\titlerunning{An Architecture for Deploying RL in Industrial Environments}

%
%
\author{Georg Schäfer \and Reuf Kozlica \and Stefan Wegenkittl \and Stefan Huber}
\authorrunning{G. Schäfer et al.}
%
\institute{Salzburg University of Applied Sciences, Salzburg, Austria \\
	\email{\{georg.schaefer, reuf.kozlica, stefan.wegenkittl, stefan.huber\}@fh-salzburg.ac.at}}
\maketitle              

\begin{abstract}
Industry 4.0 is driven by demands like shorter time-to-market, mass customization of products, and batch size one production. \gls{rl}, a machine learning paradigm shown to possess a great potential in improving and surpassing human level performance in numerous complex tasks, allows coping with the mentioned demands. In this paper, we present an OPC~UA based \gls{ot}-aware \gls{rl} architecture, which extends the standard \gls{rl} setting, combining it with the setting of digital twins. Moreover, we define an OPC~UA information model allowing for a generalized plug-and-play like approach for exchanging the \gls{rl} agent used. In conclusion, we demonstrate and evaluate the architecture, by creating a proof of concept. By means of solving a toy example, we show that this architecture can be used to determine the optimal policy using a real control system.

\end{abstract}
\section{Motivation}
\label{sec:motivation}
In addition to supervised and unsupervised learning, \gls{rl} represents a third large machine learning paradigm. In contrast to supervised and unsupervised learning, particular skills and knowledge are gained by an agent through an extensive trial and error process. Since it showed a potential of surpassing human level performance in various complex tasks \cite{nian2020rlreview}, and without the need of pre-generated data in advance, \gls{rl} is suited for applications in an industrial environment. 

Through interaction with an environment, an agent can identify an optimal policy to autonomously execute complex control tasks \cite{kober2013reinforcement}. Depending on the reward definition, productivity can be maximized, thereby lowering the cost factor of the production.
A \gls{rl} agent is able to generate new data through exploration of its environment, helping to cope with lack of data typical for \gls{ot} environments.
Additionally, the agent is exploiting its environment, thus leading to early detection of unexpected behavior in simulations. 
This knowledge can be used to implement more realistic digital representations of the environment \cite{kober2013reinforcement}, supporting the creation of an industrial plant through different stages of its life cycle. Moreover, it has been shown to possess great potential for process optimization in \glspl{ics}, allowing to cope with demands coming along with Industry~4.0, including: shorter time-to-market, mass customization of products, and batch size one production, cf.~\cite{nian2020reviewrl}.

Although \gls{rl} is a promising machine learning paradigm, there are still quite a few challenges to be overcome. One of those challenges is the deployment and integration of \gls{rl} models into \gls{ot}.

\section{Introduction to Reinforcement Learning}
\label{sec:reinforcement_learning}
\gls{rl} aims to imitate the natural human learning behavior, especially in the early stage of human life, where exploration plays an important role. Thus, it is a computational approach to learning from interaction \cite{sutton2018reinforcement}.

\begin{figure}[ht]
  \centering
  \begin{tikzpicture}[%
      boxnode/.style = {draw, inner sep=1ex, fill=black!3},
      node distance=10em]
    \node[boxnode]                (agent) {RL agent};
    \node[boxnode,right of=agent] (env)   {RL environment};

    \draw[->] (agent) -- ++(0,2em)  -| node[pos=0.25,above] {action}        (env.135);
    \draw[<-] (agent) -- ++(0,-2em) -| node[pos=0.25,below] {state, reward} (env.225);
  \end{tikzpicture}

  \caption{Standard Reinforcement Learning setup.}
  \label{fig:rl_setup_basic}
\end{figure}
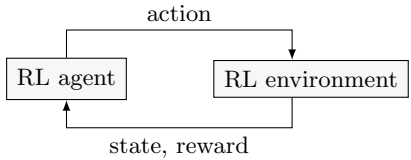

The common \gls{rl} setting is shown in \cref{fig:rl_setup_basic}. Its two main building blocks are the agent and the environment. The agent is in a specific state $s \in \mathcal{S}$ and can perform actions $a \in \mathcal{A}$, where each of those may be members of either discrete or continuous sets and can be multidimensional \cite{kober2013reinforcement}. The action $a$ performed by the agent is changing the state of the environment, and is being evaluated by the environment. For every step, the environment is sending the evaluation results in the form of a single scalar value, reward $\mathcal{R}$, to the agent. The goal of the agent is to maximize the accumulated reward over the long run. 
During the interaction with the environment, the agent is aiming to find a policy $\pi$, mapping states to actions, that maximizes the overall return \cite{kaelbling1996survey}. The policy may be either deterministic or probabilistic. In the case of a deterministic policy, the same action is always used for a given state $a = \pi(s)$, whereas the probabilistic policy maps a distribution over actions when in a specific state $a \sim \pi(s,a)$ \cite{sutton2018reinforcement}.

\section{Challenges}
\label{sec:challenges}
Integration of \gls{rl} algorithms into \gls{ot} proves to be non-trivial, since there are some \gls{ot} specific characteristics which do not fit into the regular \gls{rl} setting. In our paper, we propose an \gls{ot}-aware \gls{rl} architecture that specifically addresses the following \gls{ot} characteristics:

\begin{enumerate}
  \item \label{req:geographical_distribution}
  \textbf{Geographical distribution:} \gls{ot} systems are often located across different sites, sometimes even in remote locations, where a visit is made difficult due to travel restrictions or inhospitable site conditions. Thus, \gls{ot} systems are often geographically broadly dispersed. This fact impairs the regular \gls{rl} setting, where the agent typically interacts with local environments.
  
  \item \label{req:platform_independence}
  \textbf{Platform independence:} \gls{ot} systems are of a heterogeneous nature, many of them using proprietary technology, concerning hardware platform, operating system, or network protocols. On such systems, state of the art machine learning algorithms including \gls{rl} models are hard to be integrated. This makes the integration of a \gls{rl} agent into \gls{ot} systems built by different manufacturers difficult.
\end{enumerate}

Besides the above-mentioned \gls{ot}-specific characteristics which impair the typical \gls{rl} setting, we also consider the following requirements for the sake of general applicability of the architecture. This way, our architecture also addresses the needs of a framework for:

\begin{enumerate}
  \setcounter{enumi}{2}
  \item \label{req:rl_agent_agnosticism}
  \textbf{\gls{rl} agent agnosticism:} 
  Being able to rely on generally available \gls{rl} agent implementations\footnote{E.g., \url{https://www.tensorflow.org/agents}} is important to reduce development effort. Therefore, the proposed architecture enables a plug-and-play like usage of different \gls{rl} agents through definition of an OPC~UA information model in \cref{sec:opc_ua_information_model}.
  \item \label{req:digital_representation}
  \textbf{Digital representation:} 
  A major challenge when working with \gls{rl} algorithms is the sample inefficiency: an enormous amount of interaction data is required, which makes training expensive~\cite{mai2022sampleefficientldrl}. Moreover, this can result in unbearable cost for real-world applications. Our architecture addresses this issue through enabling the usage of digital representations in the initial training of the \gls{rl} agent and seamless transition to the real system. This can lead to a significant reduction of overall training time and help to cope with real-time constraints of real control systems and the lack of real-world samples~\cite{kober2013reinforcement}.
\end{enumerate}

\section{Proposed Architecture}
\label{architecture}

\begin{figure}[ht]
  \centering
  \begin{tikzpicture}[%
      boxnode/.style = {draw, inner sep=1ex, fill=black!3},
      node distance=10em]
    \node[boxnode]                (agent) {RL agent};
    \node[boxnode,right of=agent] (env)   {RL mapper};
    \node[boxnode,right of=env]   (phy)   {OPC~UA nodes};

    \draw[->] (agent) -- ++(0,2em)  -| node[pos=0.25,above] {action}        (env.135);
    \draw[<-] (phy)   -- ++(0,2em)  -| node[pos=0.25,above] {actuation}     (env.45);
    \draw[<-] (agent) -- ++(0,-2em) -| node[pos=0.25,below] {state, reward} (env.225);
    \draw[->] (phy)   -- ++(0,-2em) -| node[pos=0.25,below] {sensing}       (env.-45);
  \end{tikzpicture}

  \caption{Interaction between a RL agent and OPC~UA nodes using a RL environment.}
  \label{fig:rl_setup}
\end{figure}
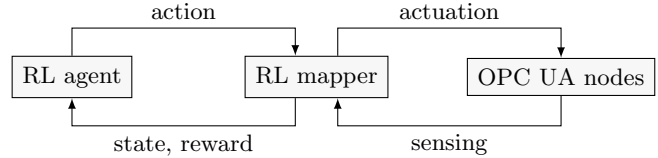

Facing the requirements defined in \cref{sec:challenges}, we propose an \gls{ot}-aware OPC~UA-based architecture.
\cref{fig:rl_setup} gives a simplified overview. Essentially, the standard \gls{rl} setting~\cite{sutton2018reinforcement} is extended by ``OPC~UA nodes''. The nodes can be aggregated using $n$ OPC~UA servers, each of them described as a finite set $\mathcal{O}_i$, containing all nodes of a specific server, where $\mathcal{O}=\mathcal{O}_1 \times \mathcal{O}_2 \times ... \times \mathcal{O}_n$. In a certain sense, we combine the \gls{rl} setting presented in \cref{sec:reinforcement_learning} with that of digital twins as defined by \cite{kritzinger2018digital}.

With OPC~UA, the communication can either occur via an optimized binary TCP protocol or using firewall-friendly Web Services by standards like SOAP and HTTP~\cite{mahnke2009opc}, fulfilling requirement~\ref{req:geographical_distribution}. Regardless of the manufacturer, every TCP/IP enabled \gls{ot} platform can implement OPC~UA, hence meeting the \nth{2} requirement. OPC~UA supports not only real hardware but also different types of simulations, thus allowing the architecture to fulfill requirement~\ref{req:digital_representation}.

The \gls{rl} action and state space is in a relationship with the OPC~UA address space, and the corresponding translation is performed by the \gls{rl} mapper. In particular, each agent's action $a \in \mathcal{A}$ is turned into an OPC~UA call\footnote{In this context an
OPC~UA call is a procedure call, which may contain a series of computational steps to be carried out.} using the function $f$ in \eqref{eq:map_action_to_opc_ua}. This function sets the corresponding actuators using the OPC~UA client-server model.

\begin{equation}
    \label{eq:map_action_to_opc_ua}
    f: \mathcal{A} \rightarrow \mathcal{O}
\end{equation}

 Vice versa, the environment is notified using the OPC~UA PubSub~\cite{opcua14} on each sensor change, mapping OPC~UA sensor nodes to states using the function~$g$:

\begin{equation}
    \label{eq:map_opc_ua_to_state}
    g: \mathcal{O} \rightarrow \mathcal{S}
\end{equation}

After each sensor change, a reward evaluation is triggered and a state space transition occurs, which is forwarded to the agent. By preserving the standard \gls{rl} setting, the agent can easily be substituted, fulfilling requirement~\ref{req:rl_agent_agnosticism}.

\subsection{Mapping Automation using OPC~UA Information Modelling}
\label{sec:opc_ua_information_model}
OPC~UA is not only used for data transport, but also for information modelling. The information model defines the address space of an OPC~UA server, which is used to navigate through the OPC~UA nodes, providing information about the available data \cite{opcua5}. Throughout this section, we will use technical terms of the information modelling mechanisms of OPC~UA as introduced by \cite{mahnke2009opc}.

By creating custom \textit{ObjectTypes} for nodes which should be accessible for the \gls{rl} agent, the mapping from \eqref{eq:map_action_to_opc_ua} and \eqref{eq:map_opc_ua_to_state} can be automated. Therefore, multiple custom \textit{ObjectTypes}, like \textit{IntObservation}, \textit{DoubleObservation}, \textit{IntAction} and \textit{DoubleAction}, have been created. These \textit{ObjectTypes} require three variables with a corresponding \textit{DataType} (e.g. \textit{Int32} and \textit{Double}):

\begin{enumerate}
    \item \textbf{min}: The smallest possible value that can be assigned to a node, including the value specified.
    \item \textbf{max}: The largest possible value that can be assigned to a node, including the value specified.
    \item \textbf{step}: The step-size of the value, allowing a discretization.
\end{enumerate}

By using a \textit{HasProperty} reference, existing nodes having a \textit{BaseDataVariableType} as \textit{TypeDefinition}, e.g. variables, can be extended to automatically provide the \gls{rl} agent with the required information, answering the following questions:

\begin{itemize}
    \item Is the node relevant for the \gls{rl} setting?
    \item Is the node used to define the action space $\mathcal{A}$?
    \item Is the node used to define the observation space $\mathcal{S}$?
    \item What are all the possible values the node might have?
\end{itemize}

Algorithm \ref{lst:defining_action_and_observation_spaces} shows how the required mapping $f$ and $g$ in \eqref{eq:map_action_to_opc_ua} and \eqref{eq:map_opc_ua_to_state} can be automated using pure information modelling.

\begin{algorithm} \caption{Determining the action and observation space} \label{lst:defining_action_and_observation_spaces}
\begin{algorithmic}[1]
\Require $\mathcal{O}$, the set containing all OPC~UA nodes
\State Initialize $\mathcal{A} \gets \varnothing$, $\mathcal{S} \gets \varnothing$, $ActionSets \gets \varnothing$ and $StateSets \gets \varnothing$

\ForEach {$o \in \mathcal{O} $}
    \If{$o$ has property \textit{ActionNode}}
        \State $ActionSets$.append(min($o$):step($o$):max($o$))
    \EndIf
    \If{$o$ has property \textit{ObservationNode}}
        \State $ObservationSets$.append(min($o$):step($o$):max($o$))
    \EndIf
\EndFor

\ForEach {$actionSet \in ActionSets $}
    \State $\mathcal{A} \gets \mathcal{A} \times actionSet$
\EndFor

\ForEach {$observationSet \in ObservationSets $}
    \State $\mathcal{S} \gets \mathcal{S} \times observationSet$
\EndFor

\end{algorithmic}
\end{algorithm}

\section{Proof of Concept}
\begin{figure}[htp] \centering{
\includegraphics[scale=1.0]{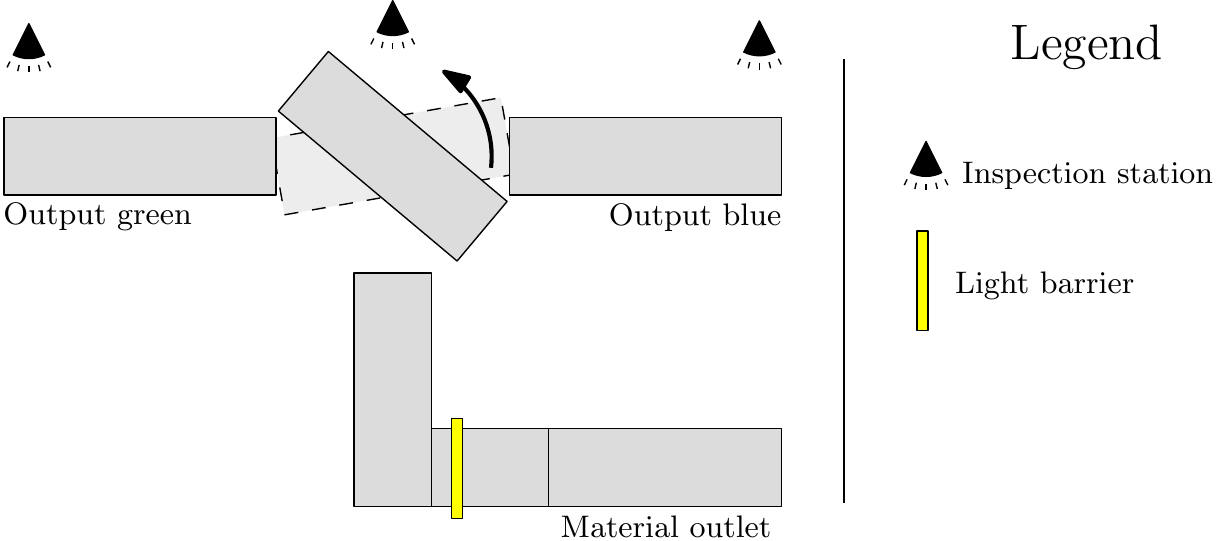}}
\caption{Structure of the material flow plant}
\label{fig:structure_poc}
\end{figure}

To demonstrate and evaluate the proposed architecture, a proof of concept has been created. A sorting task on a material flow plant was chosen to be solved. Either a green or blue material may be released at the material outlet. Afterwards, it should be transported using the conveyor belts as well as the turntable to either the left or right side, depending on the color of the material, as shown in \cref{fig:structure_poc}.

In order to evaluate the integration of a real control system, a \gls{hil} simulation~\cite{vdi3693} was used. Therefore, the integration of a real control system could be evaluated.

To not only show the deployment of a learned policy on a real control system, but also the learning of it within a reasonable time, the complexity regarding the action and state space $\mathcal{A}$ and $\mathcal{S}$ has been reduced to a minimum. Therefore, the action space $\mathcal{A}$ consists of four actions, determined by two boolean actuators to control the direction as well as the rotation of the turntable. The observation space $\mathcal{S}$ is determined by two sensors: one boolean light barrier and a color inspection station returning one of three possible values for the color green, the color blue as well as no detected material. This results in a total of six possible states, with two invalid states as long as there is no more than one material simultaneously on the material flow plant.

\begin{figure}[ht] \centering{
\includegraphics[scale=1.0]{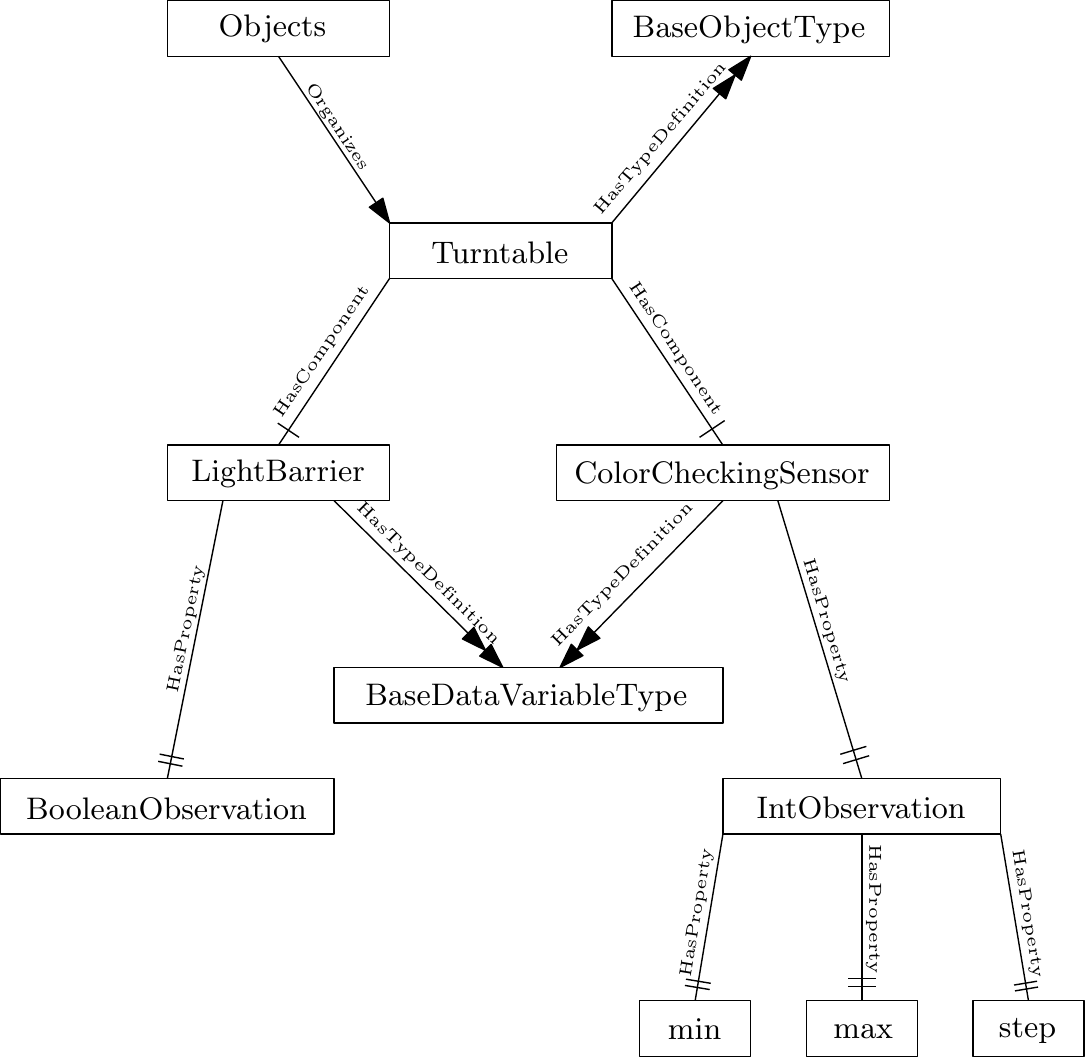}}
\caption{An excerpt of the OPC~UA information model defining the state space for the proposed sorting task.}
\label{fig:opcua_information_model_for_sorting_task}
\end{figure}

\cref{fig:opcua_information_model_for_sorting_task} shows an excerpt of the OPC~UA information model for the proposed sorting task, illustrating the used nodes comprising the state space $\mathcal{S}$. The \textit{Turntable} node is organized by the global \textit{Objects} node, and it has an arbitrary amount of components. Each component having a \textit{BaseDataVariableType} may be extended with a property introduced in \cref{sec:opc_ua_information_model} using a \textit{HasProperty} reference. Using these properties enables an automatic generation of the action and state space $\mathcal{A}$ and $\mathcal{S}$ required by the \gls{rl} agent by pure modelling.

To determine the reward $r \in \mathcal{R}$ additional sensors are needed. On each end of the material flow plant, a color inspection station is integrated. Additionally, a light grid is attached to the bottom of the plant, enabling a detection of dropped materials. To determine if a material got stuck, the passed time is logged and evaluated. This results in four possible rewards:

\begin{equation}
r =
\begin{cases}
+5 & \text{if material is transported to the correct station} \\
-1 & \text{if material is transported to the wrong station} \\
-3 & \text{if material is dropped} \\
-5 & \text{if material gets stuck}
\end{cases}
\label{eq:reward}
\end{equation}

Reducing the task to its bare minimum enables an identification of the optimal policy with a variety of \gls{rl} agents. For this example, Q-learning~\cite{watkins1989learning} with a constant learning rate of $\alpha = 0.4$ and a discount factor of $\gamma = 0.9$ was chosen. To allow for exploration, a $\epsilon$-greedy exploration strategy with $\epsilon = 0.1$ was used.
The optimal policy could be identified within 80 minutes using a total of 150 episodes. 

\section{Conclusion and Discussion}
Integrating \gls{rl} into \gls{ot} systems turns out to be challenging because of the characteristics of both worlds. To tackle these challenges, an \gls{ot}-aware \gls{rl} architecture has been proposed. In a proof of concept, we have demonstrated that an optimal policy was learned by the agent for a toy task, and we showed that the integration of a real programmable logic controller relying on real communication is possible. 
Using the proposed architecture in a proof of concept, we showed that the challenges mentioned in \cref{sec:challenges} can be tackled. 
We have demonstrated that an optimal policy was learned by the agent for this task, and we showed that the integration of a real programmable logic controller relying on real communication is possible.
Nevertheless, to support the development of an industrial system throughout the whole life cycle, we need to use different kinds of simulations, depending on the current development stage. 
To simplify the creation of industrial systems, we plan to create different simulations \cite{vdi3693}, and different digital representations~\cite{kritzinger2018digital}.

%
%
%
\bibliographystyle{splncs04}
\bibliography{references}
\end{document}